\pdfoutput=1
\documentclass[a4paper, 10 pt, conference]{ieeeconf}

\IEEEoverridecommandlockouts                              

\usepackage[hidelinks]{hyperref}
\usepackage{graphics} 
\usepackage{graphicx}
\usepackage{caption}
\usepackage{algpseudocode}
\usepackage{algorithm}
\usepackage[]{units}
\usepackage{pgfplots}
\usepackage{tikz}
\usetikzlibrary{positioning}
\usetikzlibrary{shapes.geometric}
\usetikzlibrary{shapes.misc}
\usetikzlibrary{external}
\usetikzlibrary{arrows}

\usepackage{paralist}
\usepackage{mathtools}

\usepackage{xcolor}
\usepackage{siunitx}
\usepackage{multirow}
\usepackage{epstopdf}
\usepackage{todonotes}
\graphicspath{ {Pictures/} }
\usepackage{epstopdf}
\usepackage{tikz}
\usetikzlibrary{shapes,arrows,patterns}
\usetikzlibrary{arrows.meta}
\usetikzlibrary{plotmarks}
\usetikzlibrary{calc,decorations.markings,math}
  \usepgfplotslibrary{patchplots}
\usepackage{epsfig}
\usepackage{float}
\usepackage{amsmath} 
\usepackage{amssymb}  
\usepackage{mathtools}
\graphicspath{ {Pictures/} }
\DeclareGraphicsExtensions{.pdf,.eps}
\usepackage[acronym,nomain,nonumberlist]{glossaries}
\newacronym{mav}{MAV}{Micro Aerial Vehicle}
\newacronym{dl}{DL}{Deep Learning}
\newacronym{nmpc}{NMPC}{Nonlinear Model Predictive Control}
\newacronym{imu}{IMU}{Inertial Measurement Unit}
\newacronym{cnn}{CNN}{Convolutional Neural Network}
\newacronym{relu}{ReLU}{Rectified Linear Unit}
\newacronym{mlp}{MLP}{MultiLayer Perceptron}
\newacronym{mpc}{MPC}{Model Predictive Control}
\newacronym{nn}{NN}{Nueral Network}
\newacronym{gnss}{GNSS}{Global Navigation Satellite System}
\newacronym{sfm}{SfM}{Structure from Motion}
\newacronym{ar}{AR}{Augmented Reality}
\newacronym{ros}{ROS}{Robot Operating System}
\newacronym{panoc}{PANOC}{Proximal Averaged Newton-type method for Optimal Control}
\newacronym{gps}{GPS}{Global Positioning System}
\newacronym{vi}{VI}{Visual Inertia}
\newacronym{fps}{fps}{Frame Per Second}
\newacronym{mae}{MAE}{mean absolute error}
\newacronym{dof}{DoF}{Degree of Freedom}
\newacronym{ekf}{EKF}{Extended Kalman Filter}
\newacronym{ahe}{AHE}{Adaptive Histogram Equalization}
\newacronym{sbc}{SBC}{Single Board Computer}
\newacronym{pwm}{PWM}{Pulse Width Modulation}
\newacronym{esc}{ESC}{Electronic Speed Controller}
\newacronym{fcu}{FCU}{Flight Controller Unit}
\newacronym{vio}{VIO}{Visual Inertia Odometry}
\newacronym{api}{API}{Application Programming Interface}
\newacronym{otg}{OTG}{One-The-Go}
\newacronym{slam}{SLAM}{Simultaneous Localization And Mapping}
\newacronym{zepa}{ZEPA}{Zero Entry Production Areas}
\pgfplotsset{compat=1.14}
\usepackage{url}

\usepackage{rotating}

  \usepackage{amssymb}
\usepackage{pifont}
\newcommand{\xmark}{\ding{53}}

\title{\LARGE \bf MAV Development Towards Navigation in Unknown and Dark Mining Tunnels*\thanks{*This work has been partially funded by the European Unions Horizon 2020 Research and Innovation Programme under the Grant Agreement No. 730302 SIMS.}}

\author{Dariusz Kominiak, Sina Sharif Mansouri, Christoforos Kanellakis, and George Nikolakopoulos \thanks{Robotics Team, Department of Computer, Electrical and Space Engineering, Lule\r{a} University of Technology, Lule\r{a} SE-97187, Sweden,  Emails:\texttt{\{darkom, sinsha, chrkan, geonik\}@ltu.se}}
}

\begin{document}
\maketitle
\thispagestyle{empty}
\pagestyle{empty}

\begin{abstract}
The Mining industry considers the deployment of \glspl{mav} for autonomous inspection of tunnels and shafts to increase safety and productivity. However, mines are challenging and harsh environments that have a direct effect on the degradation of high-end and expensive utilized components over time. Inspired by this effect, this article presents a low cost and modular platform for designing a fully autonomous navigating \glspl{mav} without requiring any prior information from the surrounding environment. The design of the proposed aerial vehicle can be considered as a consumable platform that can be instantly replaced in case of damage or defect, thus comes into agreement with the vision of mining companies for utilizing stable aerial robots with reasonable cost. In the proposed design, the operator has access to all on-board data, thus increasing the overall customization of the design and the execution of the mine inspection mission. The \glspl{mav} platform has a software core based on \gls{ros} operating on an Aaeon UP-Board, while it is equipped with a sensor suite to accomplish the autonomous navigation equally reliable when compared to high-end and expensive platforms.
\end{abstract}

\glsresetall
\section{Introduction}

Recently, deployments of \glspl{mav} are gaining increasing attention, especially in the mining industry~\cite{mansouri2020deploying}. The autonomous navigation of the \gls{mav}, equipped with a sensor suite has the ability to collect information, such as images, gas level, dust level, monitor the personnel, explore unknown and production areas, and minimize service times. At the same time, the deployments of \glspl{mav} increase the production and the overall safety in the mining industry, while reducing the overall operation costs aligned with the envisioned mine of \gls{zepa}~\cite{NIKOLAKOPOULOS201566}.

Furthermore, the harsh mining environments are characterized from a lack of illumination, narrow passages, wind gusts, dust and in general conditions that have a direct affect the performance of the platforms and in the worse case may cause failure in system components or even result to collision and crashes. Moreover, the commercially available \glspl{mav} rely on \gls{gps} or visual sensors, or both for performing a position estimation, however underground mines are \gls{gps}-denied environments and due to lack of any natural illumination and prominent visual and geometric features, the vision-based positioning methods are not reliable. Additionally, commercially available platforms provide manual or semi-autonomous flights, which require a pilot with a direct line of sight to the platform, a case which cannot be guaranteed in dark tunnels with multiple turns and crosses. 

\begin{figure}[htbp]
  \centering
    \includegraphics[width=1\linewidth]{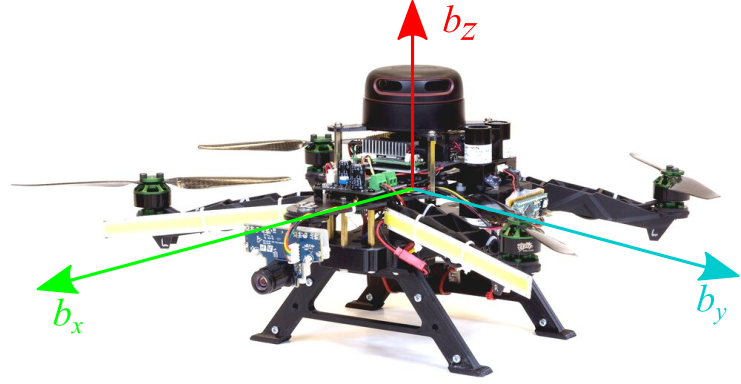}
      \caption{The proposed low-cost \glspl{mav} platform with attached body fixed frame $\mathbb{B}$.}
        \label{fig:quadcopter}
\end{figure}

The main objective of this article is to propose a low cost and modular \gls{mav} platform, as depicted in Figure~\ref{fig:quadcopter}, for autonomous navigation in dark tunnels. The platform is equipped with a 2D lidar, a single beam laser range finder, LED light bars, a PX4 optical flow, a forward looking camera, a flight controller and an on-board computer, while the software architecture is developed based on the equipped sensor suites to establish fully autonomous navigation. The proposed configuration of the \gls{mav} has been specifically designed for a direct deployed in underground mines, without a natural illumination and by that demonstrating the capability for fully autonomous \glspl{mav} navigation in such environments. 

Finally, this article discusses all the needed components in order to enable further hardware developments towards the autonomous navigation in dark tunnels. Although this work showcases the platform in tunnel navigation, the system can be deployed in similar missions including subterranean exploration~\cite{rogers2017distributed}, or Mars canyon exploration~\cite{matthaei2013swarm}.

The rest of the article is structured as it follows. Initially, the state of the art of \gls{mav} development is presented in Section~\ref{sec:relatedworks}, followed by the platform architecture in Section~\ref{sec:platformartichecture}, which discusses the corresponding hardware and software components. In Section~\ref{sec:platformperformance} the performance of the proposed platform is evaluated in an underground mine tunnel and finally in Section~\ref{Conclusions} a summary of the findings is provided.
\section{Related Works} \label{sec:relatedworks}
%

Nowadays, industry and academic sectors develop \glspl{mav} for different applications. The majority of the developed platforms by commercial companies do not provide access to raw sensor measurements or tuning parameters of the controllers. Additionally, these platforms do not allow for hardware modifications and their performance is limited to semi-autonomous navigation especially in indoor environments. These factors limit their usage and functionality in challenging environments, such as underground mine tunnels, where the platform should be modified based on the application requirements. As an example, the commercially available quad-copter Parrot Bebop 2~\cite{parrot2016parrot} is equipped with a forward looking camera, an optical flow sensor and a sonar sensor facing down, it weights $\unit[0.5]{kg}$ and it able to provide $\unit[25]{mins}$ of flight time. The platform is \gls{ros} compatible and can be used for research or teaching purposes. However, it does not have an onboard computer and provides a WiFi link so that all the computations should be on a ground station. Moreover, the user cannot modify or add extra sensors to the system, while there is no access to the low-level control architecture and sensor measurements. Another commercial product is the DJI Matrice 100~\cite{DJI}, which is a fully customizable and programmable flight platform that can be equipped with sensors required for performing an autonomous underground navigation, however it does not allow access to the low-level controllers and the raw sensor data, thus  increasing the overall complexity for the fusion of new sensor measurements. Moreover, the basic price of the platform without the sensor suites and computing unit starts from $\unit[3300]{USD}$. Table~\ref{table:industrymav} compares commercial \glspl{mav} with the proposed platform, emphasizing on important factors such as cost, \gls{ros} compatibility, sensor measurement accessibility, etc.

{\renewcommand{\arraystretch}{1.3}
\begin{table}[htbp!]
\centering
\caption{The comparison of the existing \glspl{mav} with proposed platform.}
\label{table:industrymav}
\resizebox{\linewidth}{!}{
\begin{tabular}{cccccccc}
\hline
Platform & \rotatebox{90}{\parbox{1.5cm}{Cost}} &
\rotatebox{90}{\parbox{2cm}{Sensors for autonomy}} &
\rotatebox{90}{\parbox{2cm}{Computer unit}}  & \rotatebox{90}{\parbox{2cm}{ROS}} & \rotatebox{90}{\parbox{2cm}{Data accessibility}} & \rotatebox{90}{\parbox{2cm}{Hardware Modification}} &  \rotatebox{90}{\parbox{2cm}{Spare part availability}}  \\ \hline
\textbf{Proposed Platform}  & Low  & \checkmark  & \checkmark  & \checkmark & \checkmark & \checkmark & High \\ \hline
Intel Aero  & Low  & Moderate  &  \checkmark  & Moderate  & \checkmark  & \checkmark & Low \\ \hline
DJI Matrice 100  & High  & \xmark  & \xmark  & \checkmark  & Moderate  & \checkmark & High \\ \hline
Yuneec H520  & High  & Moderate  & \xmark  & \xmark  & \xmark  & \xmark & Moderate \\ \hline
AscTec Neo & Very high  & \xmark  & \checkmark  & \checkmark  & \checkmark  & \checkmark & Moderate \\ \hline
\end{tabular}}
\end{table}
}

There are multiple open-source platforms developed for teaching proposes, such as the CrazyFlie~\cite{giernacki2017crazyflie}, the Parrot Minidrone~\cite{MDrone} and the PiDrone~\cite{brand2018pidrone}. The CrazyFlie is a small quad-copter, which provides $\unit[4]{min}$ of flight time, without pay-load and due to its size, it cannot compensate wind-gusts. The parrot Minidrone has the same drawbacks as the CrazyFlie, while the platform is not \gls{ros} compatible, a factor that is drastically limiting its usage. The PiDrone is a \gls{ros} compatible quad-copter with an onboard Raspberry Pi that runs Python and provides a $\unit[7]{min}$ flight time. Additionally, the PiDrone provides an accessible and inexpensive platform for introducing students to robotics. The drawback of this platform is lack of proper sensor suites for dark tunnels and the corresponding limited computational power for advanced algorithms and methods such as \gls{vio}. 

Furthermore, within the related literature of \glspl{mav} in underground mining operations, few research efforts have been reported trying to address challenging tasks within the mine. In~\cite{schmid2014autonomous} a visual-inertial navigation framework has been proposed, while the system was experimentally evaluated in a real-scale tunnel environment, simulating a coal mine, where the illumination challenge was assumed solved, while the platform is based on Ascending Technologies Pelican quad-copter and the authors have performed a low-level adaptation of a commercial platform, which is a complex task. In~\cite{gohl2014towards}, a more realistic approach, compared to~\cite{schmid2014autonomous}, regarding the underground localization has been performed. The FireFly hexacopter from Ascending Technologies, equipped with a \gls{vi} and a Hokuyo URG-04LX 2D laser scanner was used and it was manually guided across a vertical mine shaft to collect data for post-processing. In~\cite{ozaslan2017autonomous}, the authors addressed the problem of estimation, control, navigation and mapping for autonomous inspection of tunnels using a DJI F550 platform equipped with a Velodyne PuckLITE lidar, four Chameleon3 cameras, a PixHawk optical flow and an Intel core i7 NUC PC. The overall approach was validated through field trials, however in this case a high-end and expensive sensor suit was utilized while flying.  
\section{Platform Architecture} \label{sec:platformartichecture}
In this article, the proposed quad-copter is designed to be inexpensive, modular, autonomous, while it provides access to all the onboard raw sensor measurements. In the sequel, the hardware and software components of the overall architectures are discussed.
\subsection{Hardware Components}
\begin{figure}[htbp]
  \centering
    \includegraphics[width=1\linewidth]{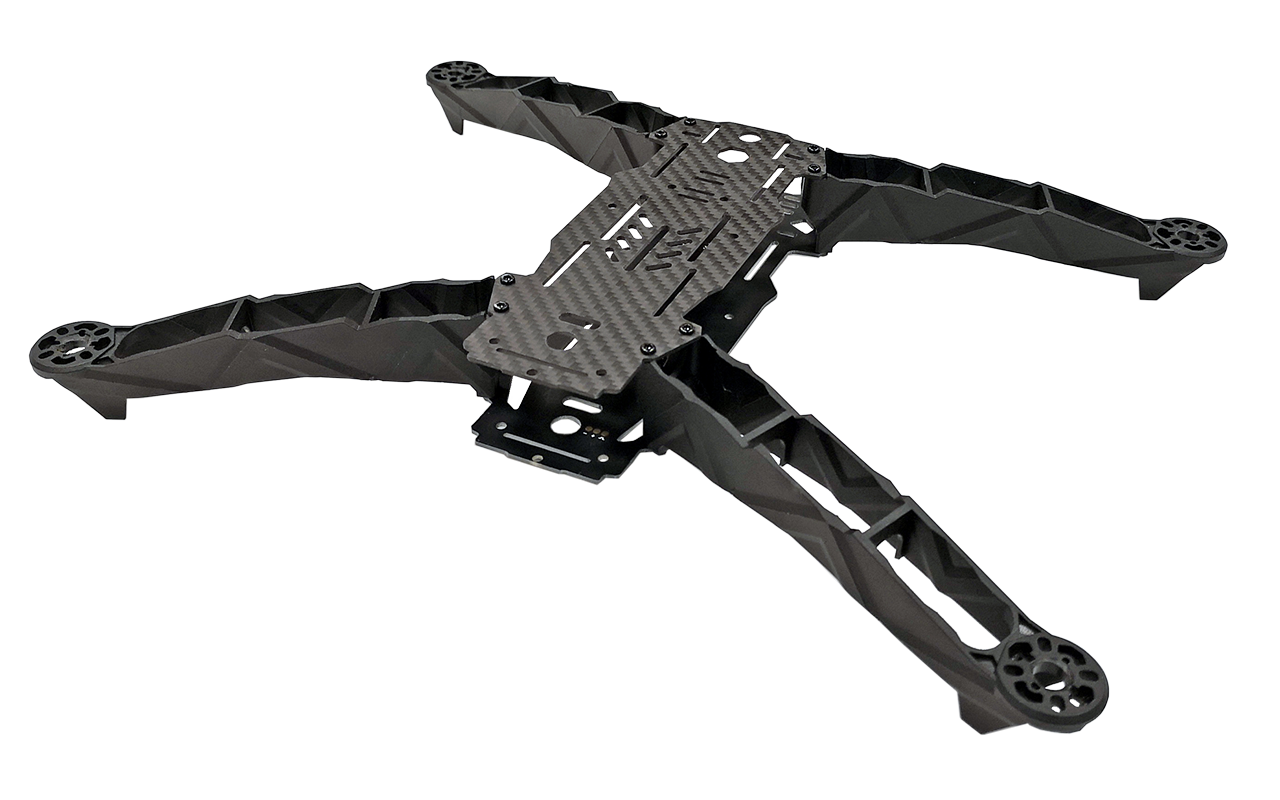}
      \caption{The quad-copter frame.}
        \label{fig:frame}
\end{figure}

\begin{figure}[htbp]
  \centering
    \includegraphics[width=0.9\linewidth]{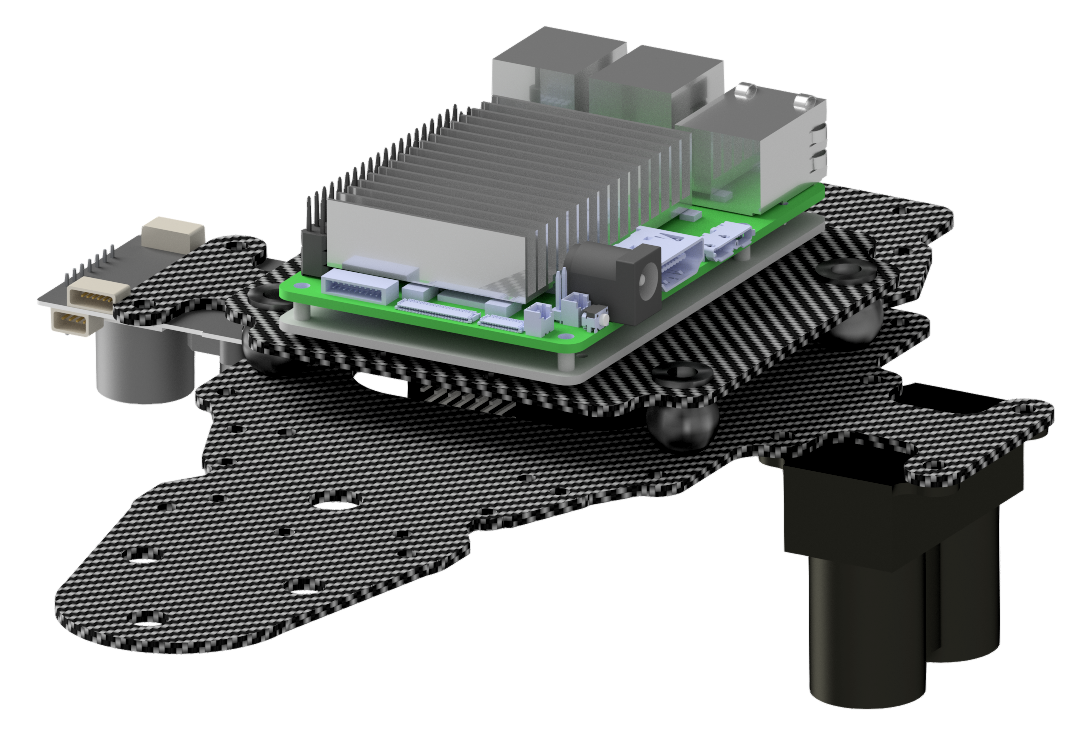}
      \caption{Modified top part of the frame with components installed}
        \label{fig:top_frame}
\end{figure}

The Enzo330 V2 330mm Wheelbase frame is selected, due to a dense market availability, low cost, durability and customizability. Figure~\ref{fig:frame} presents the corresponding frame structure. 
However, few modifications are performed to improve the frame functionality and durability as depicted in Figure~\ref{fig:top_frame}. The top part of the frame has been redesigned to allow installation of the computing unit, power modules and additional sensors. For durability reasons, the top part has been made out of carbon fiber. Additionally, extra damping for the flight controller was introduced to reduce the vast amount of vibrations generated by the motors. 

\begin{figure*}[htbp!]
  \centering
    \includegraphics[width=1\linewidth]{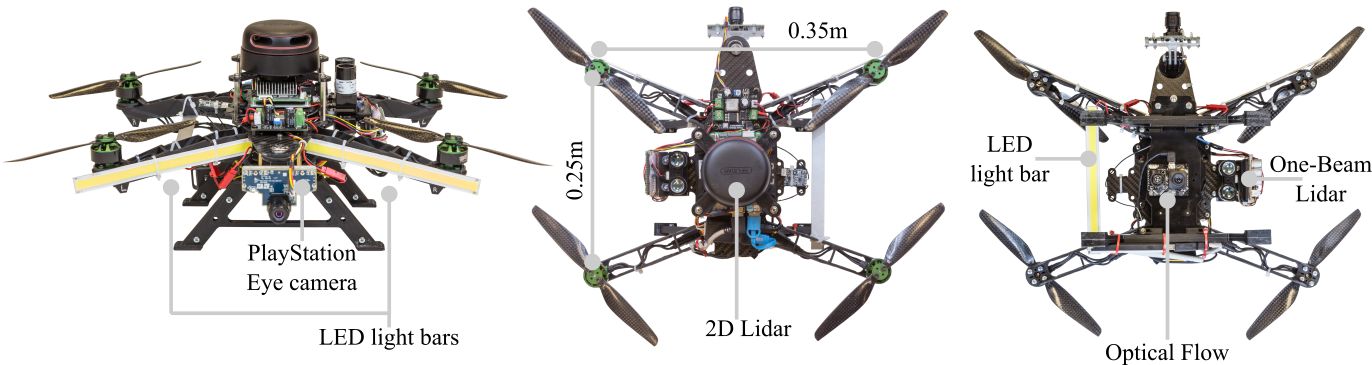}
      \caption{The developed quad-copter equipped with a forward looking camera, LED lights, an optical flow, a 2D lidar and a single beam laser range finder.}
        \label{fig:LTUplatfrom}
\end{figure*}

The landing gear has been designed and 3D printed to provide sufficient space for the sensors located on the bottom, such as an optical flow sensor and a front facing camera. Figure~\ref{fig:LTUplatfrom} depicts the modified proposed frame, equipped with sensor suits, while highlighting the dimensions of the platform. Moreover, the Multistar Elite 2308-1400 motors, carbon fiber T-Style 8x2.7 propellers and the Turnigy Multistar BLheli 32 ARM 4 in 1 32bit 31A \glspl{esc} have been selected based on the frame dimensions, estimated weight of the complete platform and the power required by the motors.

\begin{figure}[htbp!]
  \centering
    \includegraphics[width=0.9\linewidth]{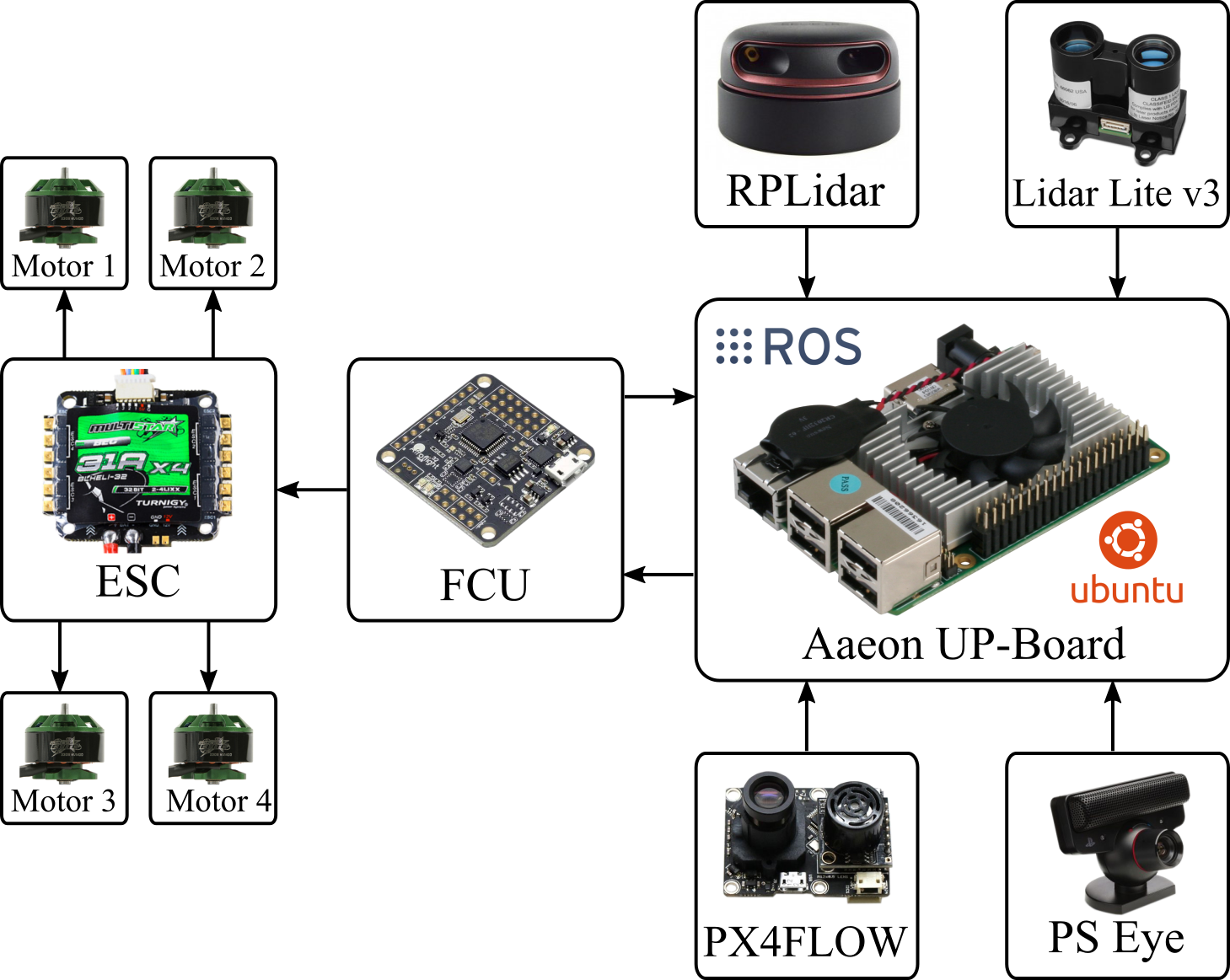}
      \caption{The proposed quad-copter hardware components.}
        \label{fig:hwmodule}
\end{figure}

Furthermore, to establish the autonomous flight, the proposed quad-copter is equipped with the hardware modules depicted in Figure~\ref{fig:hwmodule}, while Table~\ref{table:parts} provides the cost of each component and the total cost of the platform. In the following Table~\ref{table:parts} the core hardware components are discussed. 

\begin{table}[htbp!]
\caption{List of the components}
\label{table:parts}
\begin{tabular}{ccc}
\hline
\textbf{Subsystem} & \textbf{Item} & \textbf{Cost} \\ \hline
Computation & Aaeon UP-Board & \$190.00 \\
Avionics & NAZE 32 REV 6 FCU & \$35.00 \\
Avionics & 4x Multistar Elite 2308-1400 Motors & \$85.00 \\
Avionics & Turnigy Multistar 4-in-1 ESC & \$45.00 \\
Avionics & 4x 8x2.7 carbon fiber propellers & \$10.00 \\
Avionics & Enzo330 Frame with upgraded top plate & \$30.00 \\
Sensors & RPLidar A2M8 & \$280.00 \\
Sensors & LIDAR-Lite 3 & \$130.00 \\
Sensors & PX4FLOW & \$135.00 \\
Sensors & PS3 Eye Camera & \$30.00 \\
Power & Battery and DC-DC converter & \$50.00 \\
Light & LED bars and current drivers & \$75.00 \\ \hline
 & \textbf{Total cost:} & \textbf{\$1095.00} \\ \hline
\end{tabular}
\end{table}
%
\subsubsection{Flight Controller and On-board Computer}
%
The ROSFlight~\cite{jackson2016rosflight} is an embedded autopilot system that provides high-rate sensor data to \gls{ros} and a simple \gls{api} for sending commands. Thus, the AfroFlight Naze32 Rev6 has been used as \gls{fcu} in the proposed platform. Moreover, the selection of the on-board computer is based on trade-off between the cost, performance, and weight. The on-board computer should provide enough computation power to execute an autonomous navigation, a state estimation, and proper vision algorithms. Based on the requirements, the \gls{sbc} Up-Board UP-CHT01 manufactured by Aaeon was selected. The Aaeon is equipped with a Quad Core Intel Atom x5-z8350 Processor, 4GB DDR3L-1600 memory and provides six USB 2.0 and one USB 3.0 \gls{otg} ports, while it weights $\unit[195]{g}$.

\subsubsection{Sensor Suite}

The main component and cost of the platform is the sensor suite. The 2D lidar scanner, the PX4 optical flow, the PlayStation 3 Eye camera and the single beam laser range finder have been selected to provide the required information for the algorithms to establish the autonomous navigation.

The RPLidar A2M8 360 Degree Laser Scanner is the sensor that provides distance measurements of the surrounding. It uses a laser with a wavelength: $\unit[775]{nm}$ to $\unit[795]{nm}$ and provides an up to $\unit[15]{hz}$ scan frequency. The distance range is from $\unit[0.15]{m}$ to $\unit[12]{m}$, but measurements above $\unit[8]{m}$ are not reliable. Thus, above the $\unit[7]{m}$ the measurement data is not considered. In order to provide accurate altitude measurements, the LIDAR-Lite 3 Laser Rangefinder has been used as the main altitude sensor. It provides a range up to $\unit[40]{m}$ with the accuracy of $\unit[2.5]{cm}$. The sampling rate is set to $\unit[300]{hz}$. The rangefinder utilizes a laser $\unit[905]{nm}$ wavelength with a power of $\unit[1.3]{W}$. It should be highlighted that due to low temperature of the mine tunnels and vibrations of the frame, the PX4FLOW sonar sensor measurements are not reliable thus one beam range finder is used, although it has a higher cost.

The autonomous navigation in the unknown environment requires a pose estimation, however in low-illumination environments the pose estimation may not be reliable. Thus, velocity estimation is provided by the PX4FLOW optical flow and it includes a synchronized MT9V034 machine vision CMOS sensor with a global shutter, together with a L3GD20 3D Gyroscope. The sensor allows processing images at $\unit[250]{hz}$ based on the on-board 168MHz Cortex M4F micro-controller.

Finally, for collecting visual data, the PlayStation 3 Eye camera has been used that has the ability to capture a video stream with a resolution of $\unit[640 \times 480]{ pixels}$ with $\unit[60]{fps}$ or $\unit[320 \times 240]{ pixels}$ with $\unit[120]{fps}$. The camera has 56-75 degrees of horizontal field of view and the image stream can be used as \gls{ar} for human operator or vision based algorithms.

\subsubsection{Additional light sources}

Several extra light sources have been installed on the aerial platform to provide illumination for the visual sensors. Two $\unit[10]{W}$ LED bars have been installed on the front part of the quad-copters' arms and the constant current is provided by dedicated power LED drivers Recom RCD-24-0.70/W/X3. The drivers allow to modify the constant current, which will indicate a change in the light luminosity by \glspl{pwm} and analogue input signals. The measured light illumination on the $\unit[1]{m}$ distance from the \glspl{mav} with the maximum power utilized was $\unit[2200]{lux}$. Furthermore, the power LED bars and the drivers are placed under the propellers to utilize the airflow during the flight as a forced cooling. Additionally, 4 low power $\unit[10]{mm}$ LEDs have been installed on the bottom side of the \glspl{mav} for creating an optical flow sensor.

\subsubsection{Battery}
After multiple tests, the optimal battery has been selected given parameters such as size, weight, voltage and energy stored. Proposed battery, ZIPPY Compact 3300mAh 14.8V 40C 4S1P with weight of 360g provide flight time of $\unit[12]{min}$ in no wind conditions.

\subsection{Software Architecture} \label{ref:softwareartichecture}

\begin{figure}[htbp]
  \centering
    \includegraphics[width=1\linewidth]{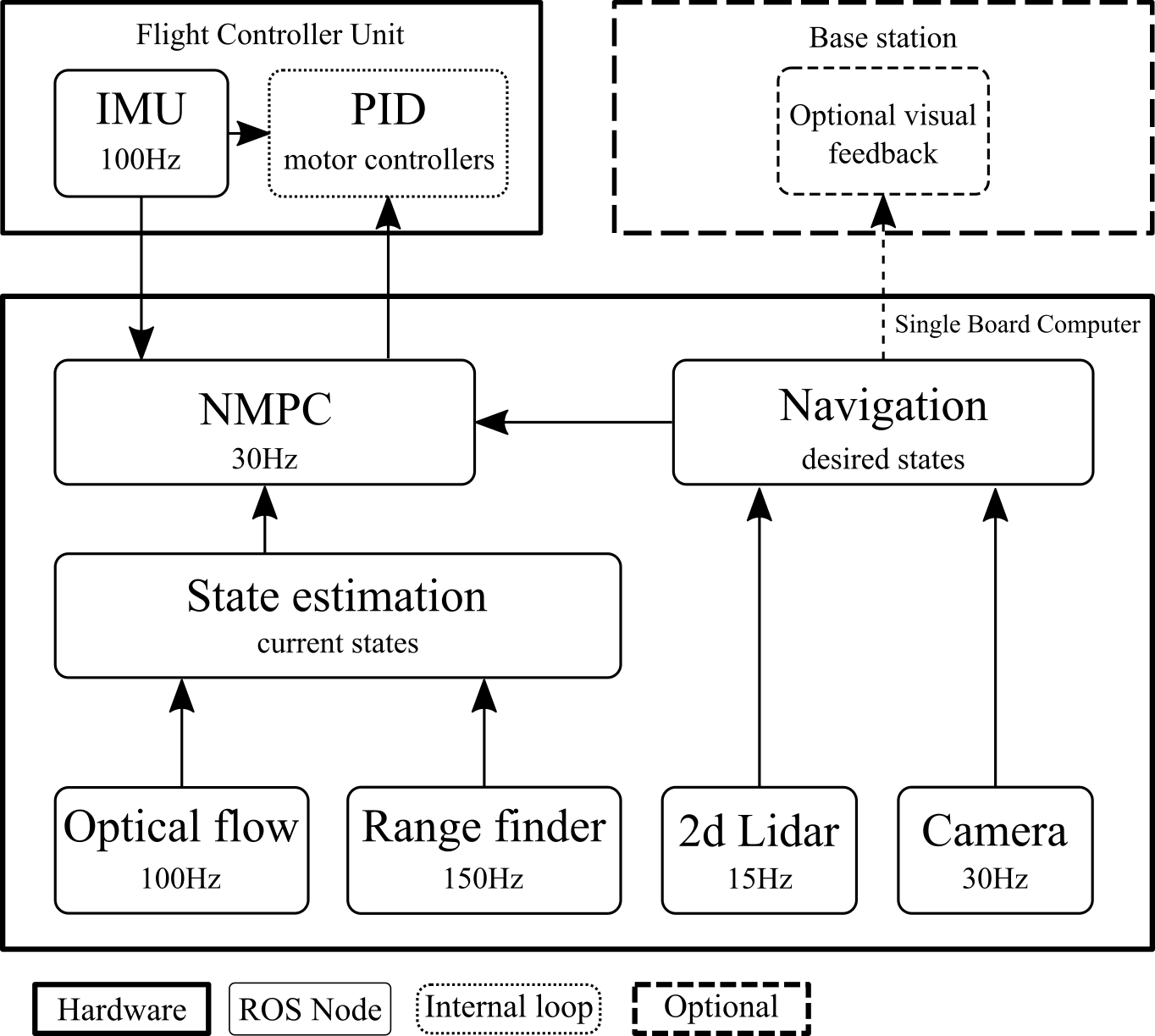} 
      \caption{The overall proposed software architecture for the navigation in the mine; For simplicity the sub-components are not shown.}
        \label{fig:schematic}
\end{figure}

The general scheme of the proposed software architecture is presented in Figure~\ref{fig:schematic}. The software architecture of the developed platform consists of the navigation, control, state estimation and visual feedback components.

 The navigation component, based on the authors previous works~\cite{mansouri2020deploying, mansouri2018dl_tunnel} and~\cite{raad2018_tunnel}, incorporates data from the on-board front facing camera or the 2D lidar for following the tunnel axis. The output of the navigation component includes the heading rate command for the \gls{mav} to correct the \gls{mav} heading towards the tunnel axis and the velocity and altitude references for controller components to hover at a fixed altitude and to avoid collision in case the platform flies close to obstacles. Furthermore, a state estimation module provides estimated values for altitude $z$, velocities along $x,y,z$ axes and attitude (roll and pitch) of the platform from the IMU, the optical flow sensor and the downward looking laser range finder measurements. It should be highlighted that an accurate estimation of the heading angle is not possible as magnetometer is not reliable, especially in an underground mine and \gls{gnss} is not available for underground areas. Moreover, the navigation commands, as well as the state estimation outcome, are sent to the \gls{nmpc} controller~\cite{small_panoc_2018} component, which generates control commands (thrust, roll, pitch, yaw-rate) for the flight controller. Finally, the visual feedback component consists of the sequential stream of the on-board camera images, which can be used for \gls{ar}.

\section{Platform Performance} \label{sec:platformperformance}
\begin{figure}[htbp]
  \centering
    \includegraphics[width=1\linewidth]{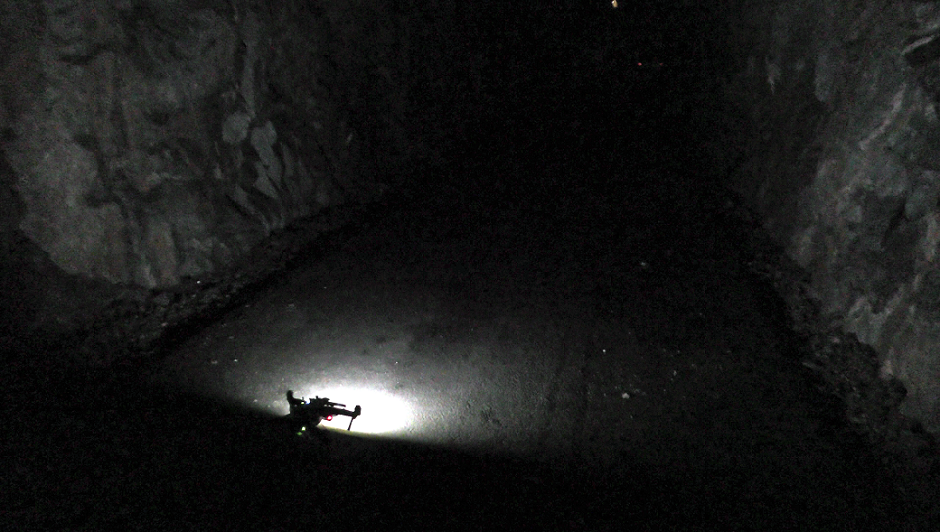} 
      \caption{Photo of a visited underground mine in Sweden.}
        \label{fig:mineboliden}
\end{figure}

This section describes the platform performance and results from each component, while the platform performs autonomous navigation in a real scale underground mine in Sweden~\cite{mansouri2019autonomous,mansouri2019vision}. Link: \url{https://youtu.be/dxMUx49a_uo} provides a video summary of the obtained results. The location of the field trials was $\unit[790]{m}$ deep, without any natural illumination sources and with a tunnel width of $\unit[6]{m}$ and a height of $\unit[4]{m}$. The area does not have strong corrupting magnetic fields, which could affect the platform sensors, while Figure~\ref{fig:mineboliden} depicts one part of the visited underground mine.

During the performed experiment, the desired altitude was selected as  $\unit[1]{m}$ with a constant velocity of $\unit[0.1]{m/s}$ aligned with the $x$-axis, while the platform is equipped with a PlayStation 3 Eye Camera and the LED light bars that provide a $\unit[460]{lux}$ illumination in $\unit[1]{m}$ distance. Moreover, a potential field method based on a 2D lidar was utilized for avoiding collisions to the tunneling walls, while measurements from a 2D lidar or a camera was used to correct the heading of the platform towards the open spaces or the tunnel axes respectively. 

\begin{figure}[htbp!]
  \centering
    \includegraphics[width=1\linewidth]{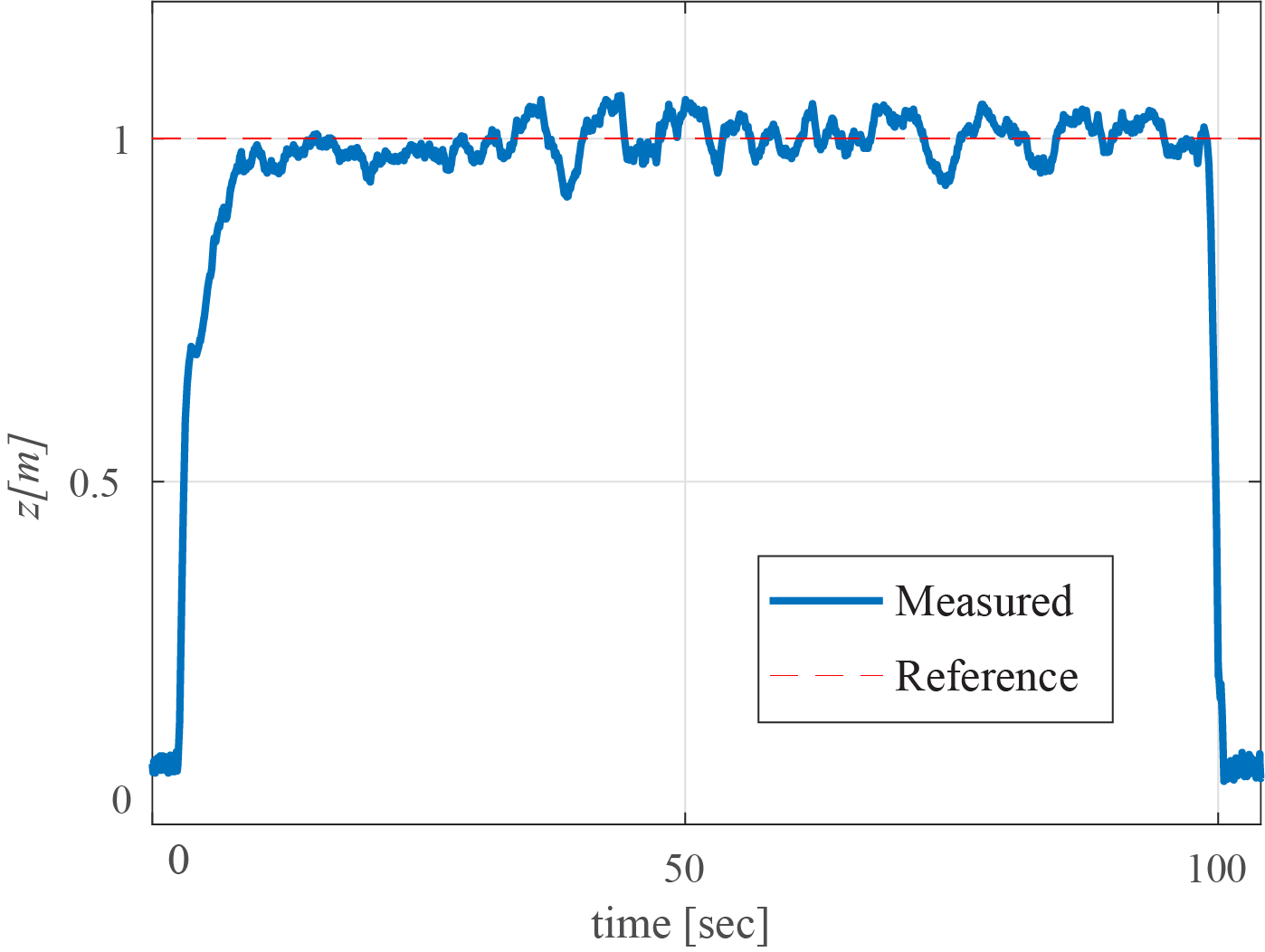}
      \caption{Altitude measurements from the downward facing range finder over time during a test flight in an underground tunnel.}
        \label{fig:altitude}
\end{figure}

The downward looking single beam laser range finder can be directly used for altitude regulation and Figure~\ref{fig:altitude} depicts the controlled achieved altitude over time for the proposed quad-copter during field trials in an underground tunnel. In this experimental case, there were no accurate height references available in the mine to evaluate the range finder measurements. However, from the overall performance of the platform, a constant altitude during the mission was successfully achieved.

\begin{figure}[htbp!]
  \centering
    \includegraphics[width=1\linewidth]{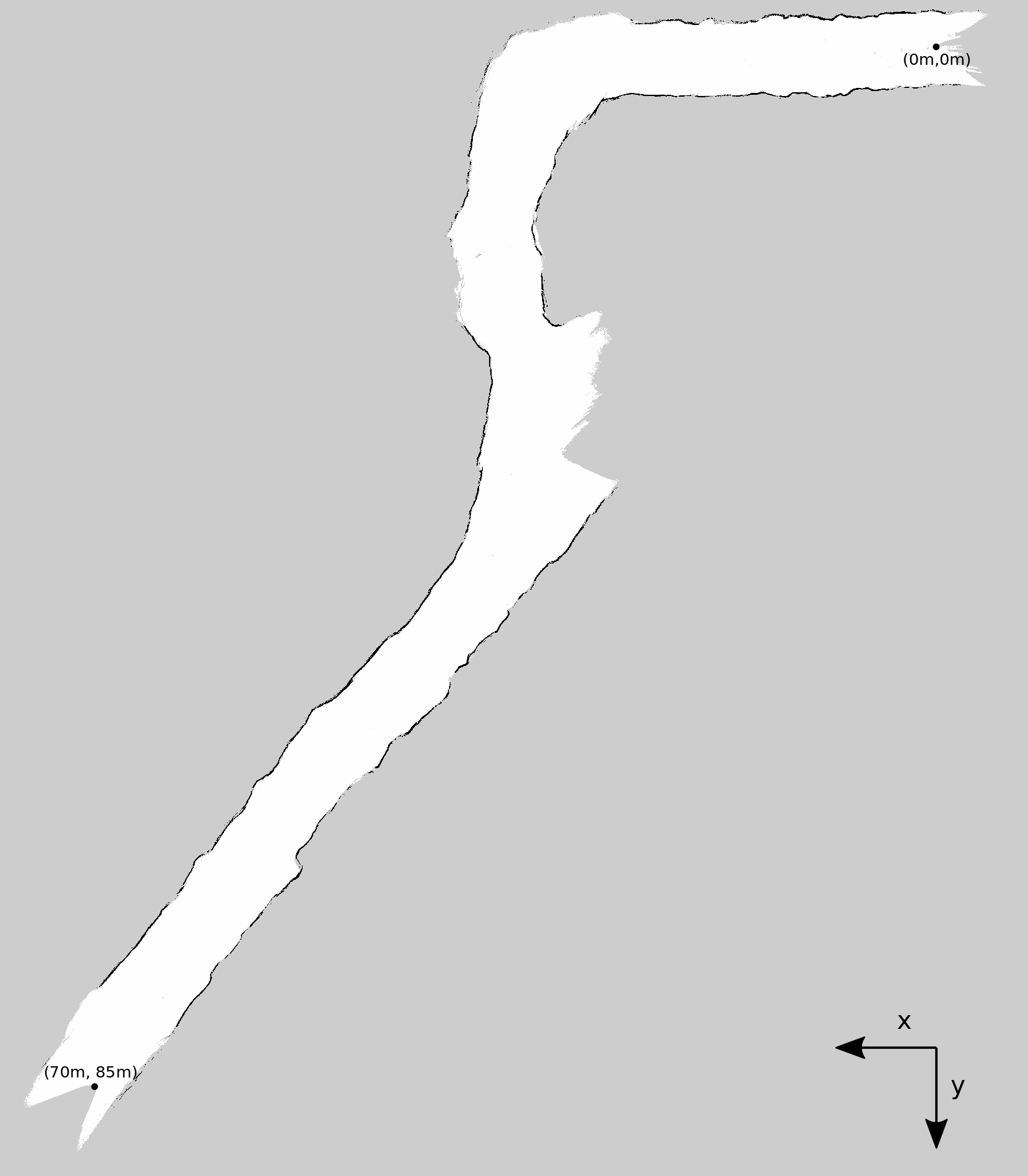}
      \caption{Obtained 2D map from laser scans while flying.}
        \label{fig:map}
\end{figure}

 The 2D laser scanner was used by algorithms for obstacle avoidance and heading correction. The collected range measurements, during the flight, can be post processed to provide a 2D map of the area. The laser scans are processed from a Lidar SLAM method~\cite{hess2016real} that is available in the ROS framework, to generate a 2D occupancy map. The map characterizes the occupied and free space of the visited underground tunnel, while it is generated online with a 1Hz update rate and a resolution of $\unit[0.05]{m/pixel}$, while the robot covered an approximate distance of (x,y)=($\unit[70]{m}$,$\unit[85]{m}$) provided from the 2D lidar processing. Figure~\ref{fig:map} depicts the 2D map of the area. The platform successfully avoids collisions from the tunnel walls and corrects its heading towards the open spaces in multiple field tests in the mine\footnote{\url{https://youtu.be/dxMUx49a_uo}}.

\begin{figure}[htbp]
  \centering
    \includegraphics[width=\linewidth]{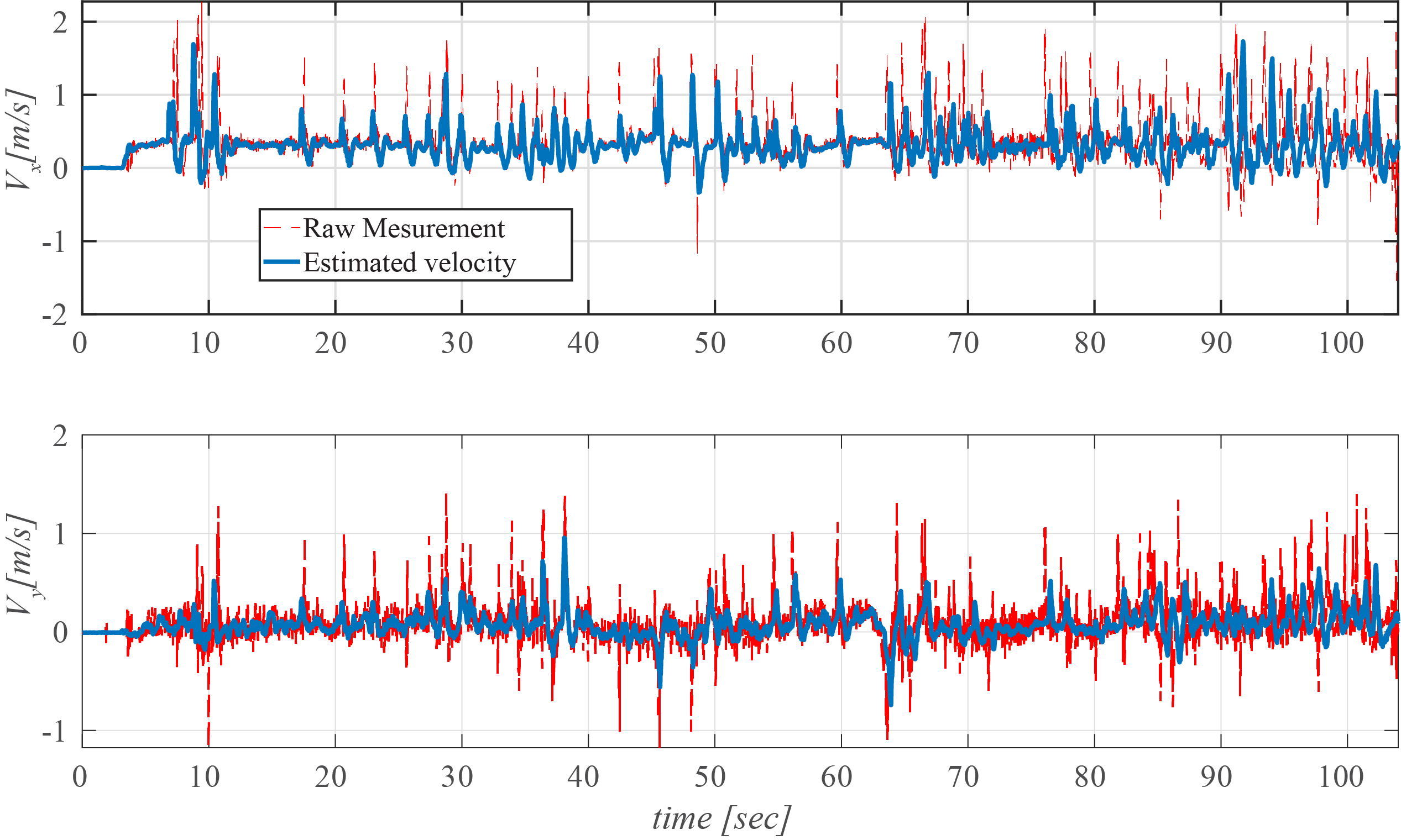}
      \caption{Time evolution of the $v_x$ and $v_y$ raw (red) and filtered (blue) corresponding velocities measurements from the downward facing optical flow sensor during a test flight in an underground tunnel.}
        \label{fig:flow}
\end{figure}

The velocity estimation for the developed system is provided from the optical flow system, however due to lack of features and illumination it is corrupted from noise measurements, thus the raw measurements were passed through low-pass filter. Figure~\ref{fig:flow} demonstrates the velocity state estimate during the field trials in the underground tunnel

\begin{figure}[htbp]
  \centering
    \includegraphics[width=\linewidth]{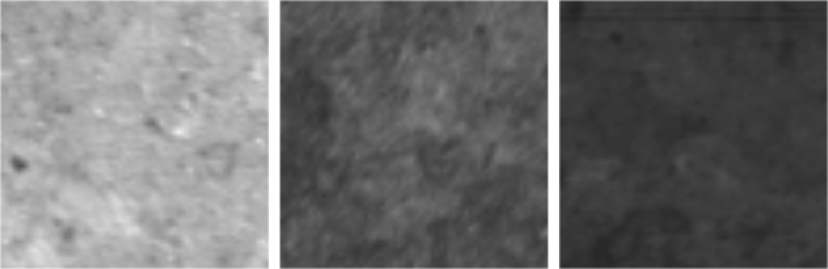}
      \caption{Example images collected from optical flow sensor.}
        \label{fig:optimg}
\end{figure}

The online information from the on-board sensors like the current altitude, \gls{mav} status, the 2D top down lidar map and the collision prediction can be overlaid to the image stream and displays to the operator as depicted in the attached video.

\section{Conclusions\label{Conclusions}} 

In this work, a low-cost high-performance quad-copter has been proposed. The developed solution is ready to fly in underground tunnels while accomplishing inspection tasks. The operator of the platform has access to all sensor measurements, as well as the information for the \gls{mav} status, assisting the algorithm development, the software implementation and the overall operation both in the preparation of the mission as well as during the mission. Finally, the developed \gls{mav} has been deployed in an unknown dark underground tunnel and successfully performs fully autonomous navigation. 
\bibliography{mybib}

\begin{thebibliography}{10}
\providecommand{\url}[1]{#1}
\csname url@samestyle\endcsname
\providecommand{\newblock}{\relax}
\providecommand{\bibinfo}[2]{#2}
\providecommand{\BIBentrySTDinterwordspacing}{\spaceskip=0pt\relax}
\providecommand{\BIBentryALTinterwordstretchfactor}{4}
\providecommand{\BIBentryALTinterwordspacing}{\spaceskip=\fontdimen2\font plus
\BIBentryALTinterwordstretchfactor\fontdimen3\font minus
  \fontdimen4\font\relax}
\providecommand{\BIBforeignlanguage}[2]{{%
\expandafter\ifx\csname l@#1\endcsname\relax
\typeout{** WARNING: IEEEtran.bst: No hyphenation pattern has been}%
\typeout{** loaded for the language `#1'. Using the pattern for}%
\typeout{** the default language instead.}%
\else
\language=\csname l@#1\endcsname
\fi
#2}}
\providecommand{\BIBdecl}{\relax}
\BIBdecl

\bibitem{mansouri2020deploying}
S.~S. Mansouri, C.~Kanellakis, D.~Kominiak, and G.~Nikolakopoulos, ``Deploying
  mavs for autonomous navigation in dark underground mine environments,''
  \emph{Robotics and Autonomous Systems}, vol. 126, p. 103472, 2020.

\bibitem{NIKOLAKOPOULOS201566}
\BIBentryALTinterwordspacing
G.~Nikolakopoulos, T.~Gustafsson, P.~Martinsson, and U.~Andersson, ``A vision
  of zero entry production areas in mines∗∗this work has been partially
  funded by the sustainable mining and innovation for the future research
  program.'' \emph{IFAC-PapersOnLine}, vol.~48, no.~17, pp. 66 -- 68, 2015, 4th
  IFAC Workshop on Mining, Mineral and Metal Processing MMM 2015. [Online].
  Available:
  \url{http://www.sciencedirect.com/science/article/pii/S2405896315019552}
\BIBentrySTDinterwordspacing

\bibitem{rogers2017distributed}
J.~G. Rogers, R.~E. Sherrill, A.~Schang, S.~L. Meadows, E.~P. Cox, B.~Byrne,
  D.~G. Baran, J.~W. Curtis, and K.~M. Brink, ``Distributed subterranean
  exploration and mapping with teams of uavs,'' in \emph{Ground/Air Multisensor
  Interoperability, Integration, and Networking for Persistent ISR VIII}, vol.
  10190.\hskip 1em plus 0.5em minus 0.4em\relax International Society for
  Optics and Photonics, 2017, p. 1019017.

\bibitem{matthaei2013swarm}
J.~Matthaei, T.~Kr{\"u}ger, S.~Nowak, and U.~Bestmann, ``Swarm exploration of
  unknown areas on mars using slam,'' in \emph{International Micro Air Vehicle
  Conference and Flight Competition (IMAV)}, 2013.

\bibitem{parrot2016parrot}
S.~Parrot, ``Parrot bebop 2,'' \emph{Retrieved from Parrot. com: http://www.
  parrot. com/products/bebop2}, 2016.

\bibitem{DJI}
\BIBentryALTinterwordspacing
 [Online]. Available: \url{http://www.test.org/doe/}
\BIBentrySTDinterwordspacing

\bibitem{giernacki2017crazyflie}
W.~Giernacki, M.~Skwierczy{\'n}ski, W.~Witwicki, P.~Wro{\'n}ski, and
  P.~Kozierski, ``{Crazyflie} 2.0 quadrotor as a platform for research and
  education in robotics and control engineering,'' in \emph{2017 22nd
  International Conference on Methods and Models in Automation and Robotics
  (MMAR)}.\hskip 1em plus 0.5em minus 0.4em\relax IEEE, 2017, pp. 37--42.

\bibitem{MDrone}
\BIBentryALTinterwordspacing
 [Online]. Available:
  \url{http://mathworks.com/hardware-support/parrot-minidrones.html}
\BIBentrySTDinterwordspacing

\bibitem{brand2018pidrone}
I.~Brand, J.~Roy, A.~Ray, J.~Oberlin, and S.~Oberlix, ``{PiDrone: An Autonomous
  Educational Drone using Raspberry Pi and Python},'' in \emph{2018 IEEE/RSJ
  International Conference on Intelligent Robots and Systems (IROS)}.\hskip 1em
  plus 0.5em minus 0.4em\relax IEEE, 2018, pp. 1--7.

\bibitem{schmid2014autonomous}
K.~Schmid, P.~Lutz, T.~Tomi{\'c}, E.~Mair, and H.~Hirschm{\"u}ller,
  ``Autonomous vision-based micro air vehicle for indoor and outdoor
  navigationgohl2014towards,'' \emph{Journal of Field Robotics}, vol.~31,
  no.~4, pp. 537--570, 2014.

\bibitem{gohl2014towards}
P.~Gohl, M.~Burri, S.~Omari, J.~Rehder, J.~Nikolic, M.~Achtelik, and
  R.~Siegwart, ``Towards autonomous mine inspectionozaslan2017autonomous,'' in
  \emph{Applied Robotics for the Power Industry (CARPI), 2014 3rd International
  Conference on}.\hskip 1em plus 0.5em minus 0.4em\relax IEEE, 2014, pp. 1--6.

\bibitem{ozaslan2017autonomous}
T.~{\"O}zaslan, G.~Loianno, J.~Keller, C.~J. Taylor, V.~Kumar, J.~M.
  Wozencraft, and T.~Hood, ``Autonomous navigation and mapping for inspection
  of penstocks and tunnels with {MAVs},'' \emph{IEEE Robotics and Automation
  Letters}, vol.~2, no.~3, pp. 1740--1747, 2017.

\bibitem{jackson2016rosflight}
J.~{Jackson}, G.~{Ellingson}, and T.~{McLain}, ``{ROSflight: A lightweight,
  inexpensive MAV research and development tool},'' in \emph{2016 International
  Conference on Unmanned Aircraft Systems (ICUAS)}, June 2016, pp. 758--762.

\bibitem{mansouri2018dl_tunnel}
S.~S. Mansouri, C.~Kanellakis, G.~Georgoulas, and G.~Nikolakopoulos, ``Towards
  mav navigation in underground mine using deep learning,'' in \emph{2018 IEEE
  International Conference on Robotics and Biomimetics (ROBIO)}.\hskip 1em plus
  0.5em minus 0.4em\relax IEEE, 2018.

\bibitem{raad2018_tunnel}
C.~Kanellakis, S.~S. Mansouri, G.~Georgoulas, and G.~Nikolakopoulos, ``Towards
  autonomous surveying of underground mine using mavs,'' in \emph{Advances in
  Service and Industrial Robotics}, N.~A. Aspragathos, P.~N. Koustoumpardis,
  and V.~C. Moulianitis, Eds.\hskip 1em plus 0.5em minus 0.4em\relax Cham:
  Springer International Publishing, 2019, pp. 173--180.

\bibitem{small_panoc_2018}
E.~Small, P.~Sopasakis, E.~Fresk, P.~Patrinos, and G.~Nikolakopoulos, ``Aerial
  navigation in obstructed environments with embedded nonlinear model
  predictive control,'' in \emph{2019 European Control Conference (ECC)}.\hskip
  1em plus 0.5em minus 0.4em\relax IEEE, 2019.

\bibitem{mansouri2019autonomous}
S.~S. Mansouri, M.~Casta{\~n}o, C.~Kanellakis, and G.~Nikolakopoulos,
  ``Autonomous mav navigation in underground mines using darkness contours
  detection,'' in \emph{International Conference on Computer Vision
  Systems}.\hskip 1em plus 0.5em minus 0.4em\relax Springer, 2019, pp.
  164--174.

\bibitem{mansouri2019vision}
S.~S. Mansouri, P.~Karvelis, C.~Kanellakis, D.~Kominiak, and G.~Nikolakopoulos,
  ``Vision-based mav navigation in underground mine using convolutional neural
  network,'' in \emph{IECON 2019-45th Annual Conference of the IEEE Industrial
  Electronics Society}, vol.~1.\hskip 1em plus 0.5em minus 0.4em\relax IEEE,
  2019, pp. 750--755.

\bibitem{hess2016real}
W.~Hess, D.~Kohler, H.~Rapp, and D.~Andor, ``Real-time loop closure in 2d lidar
  slam,'' in \emph{2016 IEEE International Conference on Robotics and
  Automation (ICRA)}.\hskip 1em plus 0.5em minus 0.4em\relax IEEE, 2016, pp.
  1271--1278.

\end{thebibliography}
\end{document}